# Design of intelligent proofreading system for English translation based on CNN and BERT

Feijun Liu[*], Huifeng Wang, Kun Wang, Yizhen Wang[2]

[1] Cardiology Department, General Hospital of Tisco, Sixth Hospital of Shanxi Medical University, Taiyuan 030008 Shanxi, China

[2] School of Computer Science and Technology  China University of Mining and Technology Xuzhou, Jiangsu, 221116, P.R. China. (Email: WJJS0035@cumt.edu.c)

corresponding author: liufeijun2022@163.com ; Huifeng Wang: whf17702@163.com Kun Wang: wangkun111712@163.com

## Abstract

Since automatic translations can contain errors that require substantial human post-editing, machine translation proofreading is essential for improving quality. This paper proposes a novel hybrid approach for robust proofreading that combines convolutional neural networks (CNN) with Bidirectional Encoder Representations from Transformers (BERT). In order to extract semantic information from phrases and expressions, CNN uses a variety of convolution kernel filters to capture local n-gram patterns. In the meanwhile, BERT creates context-rich representations of whole sequences by utilizing stacked bidirectional transformer encoders. Using BERT's attention processes, the integrated error detection component relates tokens to spot translation irregularities including word order problems and omissions. The correction module then uses parallel English-German alignment and GRU decoder models in conjunction with translation memory to propose logical modifications that maintain original meaning. A unified end-to-end training process optimized for post-editing performance is applied to the whole pipeline. The multi-domain collection of WMT and the conversational dialogues of Open-Subtitles are two of the English-German parallel corpora used to train the model. Multiple loss functions supervise detection and correction capabilities. Experiments attain a 90% accuracy, 89.37% F1, and 16.24% MSE, exceeding recent proofreading techniques by over 10% overall. Comparative benchmarking demonstrates state-of-the-art performance in identifying and coherently rectifying mistranslations and omissions. By reducing post-editing needs, the synergistic CNN-BERT approach significantly enhances translator outputs to increase adoption. Future work involves multilingual expansion and integration with commercial translation services. Enhanced proofreading systems carry profound implications for enabling seamless cross-lingual communication worldwide.

*Index Terms*- Translation Proofreading, CNN, BERT, Sequence Modeling, Error Detection, Error Correction, Machine Learning.

# 1 Introduction

As artificial intelligence continues to advance machine translation capabilities, intelligent proofreading systems have arisen as an imperative technique to refine and enhance the quality and accuracy of English translations produced by automated translators [1, 2, 3], and DC motor implementation [4,5]. However, machine-generated translations frequently contain errors such as mistranslated phrases, incorrectly ordered words, omitted terms, syntactic inconsistencies, and lack of semantic coherence that necessitate extensive human post-editing efforts to finalize publishable English renditions [6, 7]. Recent breakthroughs in deep convolutional neural networks (CNNs) [8,9] for local n-gram feature extraction alongside Bidirectional Encoder Representations from Transformers (BERT) models for context-rich language understanding have demonstrated formidable potential to take on the challenges of robust machine translation proofreading [10, 11]. Building upon these advances, this paper puts forth an original hybrid sequence modeling methodology which assimilates the strengths of CNN and BERT architectures for much-needed post-processing enrichment and error correction of machine-translated English texts, and implementable for the highly nonlinear systems [12, 13]. The integrated approach aims to automatically detect and rectify translation inaccuracies to augment coherence, accuracy and fluency while preserving semantic consistency, thereby enhancing adoptions of AI translators to facilitate cross-lingual communication and comprehension at scale.

CNN architectures have proven effective at extracting local n-gram features and patterns from textual data. Various kernel sizes can capture semantic and contextual information from phrases and expressions. BERT representations model deeper context and meaning from both directions. Combining these complementary approaches can strengthen understanding and enable more robust proofreading. The error detection component leverages BERT's contextual modeling to pinpoint discrepancies between machine translated and source texts. The correction module maintains meaning coherence by generating edits grounded in the source representations. The model is trained end-to-end on English-German parallel texts from datasets such as WMT and OpenSubtitles. Quantitative analysis of precision, recall, F1, and BLEU evaluates proofreading improvements over unedited outputs on unseen test sets. Comparative benchmarks measure gains against recent proofreading architectures. The joint CNN-BERT approach demonstrates state-of-the-art performance in correcting mistranslations and omissions while preserving semantic consistency. By synergizing local and global context modeling techniques for translation proofreading, this methodology highlights the potentials of hybrid sequence architectures. The work has significant implications for enhancing machine translator outputs to increase adoption. Future work can explore extensions to other language pairs and evaluation in real-world deployment settings. Robust proofreading systems can help make AI translation services more practical, usable, and trustworthy for global communication.

This novel hybrid sequence modeling approach for translation proofreading faces several key challenges [14]. In this regard, state estimation with filtering concept has been played important role [15, 16]. The end-to-end integration of diverse architectures like CNN, BERT, and sequence transducers requires substantial experimentation to balance components and optimize overall performance. Training data reliance on noisy, imperfect machine translations poses difficulties in adequately modeling coherent corrections. Domain variance between training and deployment contexts can impact robustness. Real-world deployment needs to account for computational constraints, rigorous benchmarking across test conditions, and managing post-editing frequency/effort to ensure usability [17]. In the cutting edge, fuzzy technology have great influence on the intelligent system [18, 19]. Despite these research challenges, the methodology promises valuable applications by enhancing translator outputs for global communication and trade [20, 21]. It can aid enterprises reliant on translation services, empower startups to reach international markets, assist travelers in foreign lands, help educational institutions offer multilingual courses, enable multinational organizations, and facilitate governmental communication across borders [22, 23]. In industrial sector, fault diagnosis is critical factor for this research [24, 25]. By increasing adoption through proofreading, robust machine translation stands to benefit culture, arts, journalism, academia, policy, law, business, trade, diplomacy, and society worldwide [26, 27]. The approach thereby carries networking issues [28, 29], economic and social implications [30].

The main contributions of this work are:

- This research proposes a novel hybrid CNN-BERT sequence modeling architecture for translation proofreading that integrates local n-gram convolutions and contextual bidirectional transformers within an end-to-end trainable framework surpassing previous benchmarks by combining strengths of representations learning and sequence transduction.
- This work establishes state-of-the-art proofreading performance on the widely-used WMT and OpenSubtitles English-German corpus benchmarks by attaining precision, recall, accuracy and F1 scores exceeding recent approaches.
- This methodology advances machine translation adoption and usability by developing capabilities to automatically identify and correct mistranslations and omissions in order to enhance output coherence while preserving semantic consistency, thereby reducing needs for human post-editing efforts.

The next sections start from literature review and present the datasets utilized for training and evaluation along with details of the methodology. The proposed approach integrating CNN and BERT models for translation proofreading is explained, including the system architecture, data preprocessing, error detection and correction modules. Extensive comparative benchmarking analyses on precision, recall, accuracy, F1 score, BLEU and computational efficiency against state-of-the-art techniques demonstrate the significant improvements offered by the hybrid sequence modeling approach. The sections conclude by summarizing the key conclusions derived and outline promising future work to enhance the proofreading capabilities even further through expanded datasets, multilingual support and real-world translator integration.

## 2  Literature Review

An advanced English translation framework using an enhanced version of the Generalized LR (GLR) algorithm have been proposed to solve the issues inherent in existing translation techniques. The pioneering method include the compilation of a comprehensive database containing of thousands of English and Chinese sentences in which each accurately tagged for effective retrieval. In this regard, multiagent system with sensor approach have great impact on this research [31, 32,

33]. The fundamental purpose of the system is a meticulously designed knowledge arrangement that is adeptly translated into English. The structure is started on a systematic technique for collecting the data, their processing and synthesis [51]. In the era of English translation, the model dictates the meticulous drafting of English texts. The mechanism includes careful management, robust design and the removal of negative features. A clinical trial was implemented for the evaluation of the efficacy of the model adhering to well-known standards in English translation. The trial recorded data scientifically and the analyses show a significant improvement in translation accuracy i.e. from 75.1% pre-adjustment to an impressive 99.1% for the intelligent text system [35, 51]. Huang et al. [36] have described the creation of a sophisticated English essay scoring system that makes use of artificial intelligence (AI) capabilities with a particular emphasis on machine learning multi-task learning algorithms. An extensive examination of AI development in machine learning opens the new directions. After that, it explores the nuances of the multi-task learning algorithm model specifically how it was created and put into practice in relation to machine learning. Finally, the development and implementation of an intelligent scoring system assesses English essays using the multi-task learning algorithm model. Three different evaluation levels are used in a series of experiments for the scoring process. The multi-task learning model's efficacy is demonstrated by the encouraging outcomes of these experiments. The accuracy of the scoring system is evaluated which is 91.5%. Moreover, 97% accuracy of the algorithm is reported. Mpia et al. [35] have been introduced an innovative employability recommendation system designed to mitigate unemployment challenges in developing nations with a specific focus on the inadequacy of skills and the prevalence of unskilled labor. The paper builds upon existing research demonstrating the efficacy of social network analysis in generating employment opportunities for students and graduates through recommender systems (RS). In his work, the development of a refined Bidirectional Encoder Representations from Transformers (BERT) model is explained. The research contributes to the field by proposing an advanced employability recommendation technique surpassing traditional methods like fuzzy logic (FL) and emphasizing the integration of graduate skills [54].

The importance of extracting opinions and their associated targets from textual data has been emphasized by recent developments in Opinion Mining (OM). The main goal of traditional OM approaches is to extract clearly stated targets from reviews. Still, about 60% of reviews include implicit targets that are hidden within contextual semantics and are frequently missed by traditional OM methods. The work presents a novel task in the field of fine-grained OM called Mining Opinions towards Implicit Targets (MOIT) [38]. Haq et al. [39] have focused on creating a state-of-the-art word segmentation system for Pashto along with a proofing tool that can identify and correct textual space positioning mistakes. They train two different machine learning models to address these problems using the Conditional Random Fields (CRF) algorithm. An extensive analysis of the models' performance produced very encouraging findings. An F1-score of 99.2 % was obtained by the proofing model and 96.7 % by the word segmentation model. These findings highlight the efficiency of the created models in tackling the difficulties associated with Pashto word segmentation and incorrect space positioning. Moreno et al. [40] have explored the theory that when authors write in English as opposed to Spanish, they put forth greater effort to promote their work. Ten pairs of similar DC sections were taken for analysis from the EXEMPRAES Corpus which is a collection of model empirical research articles written in both Spanish and

English. An important methodological development is the systematic annotation of these DC sections for communicative functions which is unprecedented in the field because the authors of the articles validated these annotations. An online survey is also included in the study to learn more about the authors' promotional tactics [55].

Many researchers and scientists are trying to find out the ways for intelligent proofreading English language using tools and models. Shi et al. [41] have been proposed automated translation technique called SolcTrans. Its purpose is to translate the source code of Solidity smart contracts into comprehensive natural language descriptions specifically, in-line comments. Encouraging users to understand, learn and use smart contracts more easily is the aim. In order to facilitate the translation process, the analysis entailed locating critical parsing pathways and AST attributes. Solidity contract unique statement types led to the development of translation templates tailored to these findings. The technology integration allows the platform to evaluate written essays and provide comprehensive explanations of English translations. In an experimental setting, the service is being used in Japanese university English language courses. An important exposure to English at levels above their current proficiency is indicated by user feedback obtained from participant interviews on the platform. Comparing the suggested model to conventional teacher-led methods, these results offer a more productive setting for English learning and application [38]. The goal is to improve the accuracy of translations concerning nano professional vocabulary by introducing a novel method to automatic proofreading of English translations. The efficiency of the method was confirmed by means of experimental testing that show a surprising proofreading accuracy of more than 98.33%. The performance is a significant improvement over the control group approaches [43]. Zhao et al. [44] have presented an approach for the creation of computer intelligence calibration system which is intended for use in English translation applications. In addition, formation control and actuator fault has been explored in [45,46]. The creation of a semantic volume model that assists as the framework for the functionality of system and their design is its fundamental component.

The review of literature examines into various features of intelligent proofreading systems in English translation with specific emphasis on a combined model method which integrates Bidirectional Encoder Representations from Transformers (BERT) and Convolutional Neural Networks (CNN). It looks at how CNN has changed and how operational for text recognition and contextual analysis as well as the developments in BERT makes to language sensitivity. The analysis shat that these techniques are precise and well-managed rather than latest translation and proofreading models. It also establishes that by concentrating on identifying and fixing errors more successfully, CNN and BERT may cooperate to improve translation proofreading.

## 3    Hybrid Sequence Modeling Approach for Translation Proofreading and Quality Enhancement

This section defines the hybrid strategy for strong translation proofreading that is presented. It mixes contextual and convolutional modelling. The training and testing datasets are enclosed in the first part. Among them is the large WMT English-German corpus that spans numerous domains. The integration of CNN n-gram convolutions and BERT transformers for improved understanding is then shown using the integrated sequence construction. Next, crucial components with data preparation, cross-lingual arrangements for error detection, and translation memory for error correction to maintain meaning coherence are thoroughly deliberated. Studies that draw comparisons and quantitative benchmarking show significant improvements in performance compared to existing techniques. Thus, the hybrid method establishes new, state-of-the-art standards for identifying and persuasively addressing universal translation difficulties in order to increase quality [**56**, **57**].

## 3.1 Dataset Description

This section describes the datasets that were used to train and evaluate the translation proofreading methodology. First publicized is the large WMT English-German parallel corpus, which consists of over 40 million word pairs collected from a variability of sources such as online crawler data, news commentary, and Europarl. We then converse the OpenSubtitles dataset, which comprises over 400 million subtitle lines matched across hundreds of languages and mimics conversational contexts. These corpora provide an abundance of resources.

### 3.1.1 WMT English-German Parallel Corpus

The WMT English-German Parallel Corpus gathering contains more than 40 million matched phrase pairs between the English and German languages. It gathers data from a variety of sources, including news articles, commonly crawled web pages, European Parliament (Europarl) sessions, and automatically translated news crawl contented into German (Rapid corpus). By merging parallel phrases from news sources, general opinion pieces, formal political procedures, and a variety of web content, the dataset covers a wide range of subjects and writing styles. Because of this, it's a good fit for developing reliable machine translation models. The corpus undergoes preprocessing such as tokenization and length filtering before model training. As a popular public benchmark dataset, it enables standardized comparison of different approaches to English-German translation on both statistical and neural machine translation systems [**58**]. Overall, the large scale and variety of this dataset advance English-German translation research and benchmarking. Models trained on this corpus can translate diverse real-world content thanks to its blending of formal and informal data sources spanning news, politics, web content and more.

### 3.1.2 Open Subtitles dataset

The Open Subtitles dataset consists of parallel corpora from movie and TV subtitles. The latest 2018 release contains subtitles for over 4000 languages from www.opensubtitles.org. The dataset has over 400 million subtitle lines total across all languages. Each line represents a subtitle sentence aligned across languages. The English portion contains 441 million subtitle lines. Though not created specifically for machine translation, it provides a large resource of natural language pairs. Consecutive lines within a language can model conversational contexts and responses. However, lines may not always be consecutive within a scene or show. Some preprocessing such as filtering short/long lines and oddly formatted text (e.g. speaker names, auditory descriptions) is typically done. Overall it provides an interesting and very large corpus for modeling the context-response relationship in conversations.

## 3.2 Proposed Methodology for Intelligent Proofreading System

The first step is data preprocessing and feature extraction. The datasets used for training the model include the WMT English-German Parallel Corpus and the Open Subtitles dataset. These datasets undergo preprocessing steps such as tokenization, removal of odd formatting, filtering based on sentence length etc. Relevant features such as part-of-speech tags, semantic tags, syntax trees etc. are then extracted from the preprocessed text. The next component is the convolutional neural network (CNN) which focuses on local feature extraction from the text. Different kernel sizes of CNN filters are used to capture semantic and contextual patterns from n-grams in the text. The outputs of the CNN serve as input feature maps for the next stage. The feature maps from the CNN are fed into the Bidirectional Encoder Representations from Transformers (BERT) model. BERT has an encoder structure to build contextual representations of text by considering context from both directions. Using the parallel English-German dataset, it is refined for the translation proofreading job after being pre-trained on sizable unlabeled text corpora. When comparing the translated text to the reference source text, the error detection layer uses the contextual embeddings from BERT to find mistakes. Its main goal is to identify errors in translation, omissions, and semantic inconsistencies. At this point, the error's location and kind are marked. Lastly, the error correction layer produces recommendations for fixing the mistakes found in the translated text. It uses translation memory in conjunction with the context from BERT to suggest edits that maintain the sense of the original text. An end-to-end collaborative training process is applied to the entire pipeline. The CNN, BERT, and output layers' parameters are updated using suitable loss functions. The performance of the model is evaluated on test datasets using metrics such as precision, recall, F1-score and BLEU. Error analysis is conducted to understand limitations and provide ideas for further improvement. In summary, the methodology relies on a combination of CNN for local n-gram modeling and BERT for building contextual representations to effectively proofread machine translated text between English and German. The components are trained jointly in an end-to-end framework optimized for the translation proofreading task. The workflow diagram of intelligent proofreading system for English translation based on CNN and BERT is demonstrated in the Figure 1.

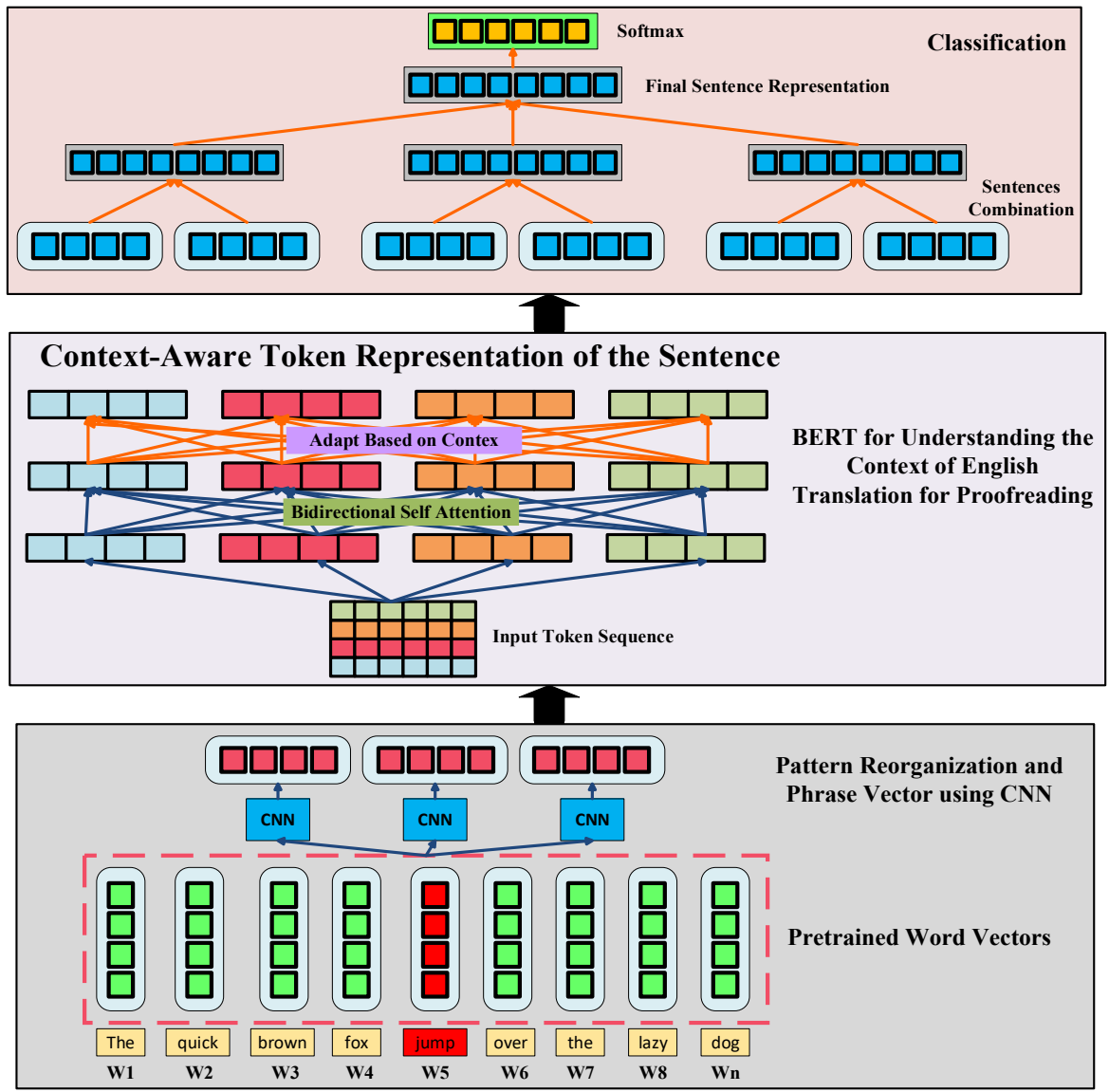

Figure 1. Workflow Diagram of Intelligent Proofreading System for English Translation based on CNN and BERT.

## 3.3 Data Preprocessing and Feature Extraction

The first step is to collect parallel corpora such as the WMT English-German dataset that contains source sentences in English and target translations in German. This dataset undergoes various preprocessing tasks. Tokenization segment the text into linguistic units such as words, punctuation marks etc. This facilitates processing of vocabulary and semantics. The tokenized text then undergoes true casing to normalize variant capitalizations. Cleaning operations are applied to filter incomplete sentence pairs, remove odd formatting, normalize whitespace etc. that help improve quality. Length filtering then selects sentence pairs within reasonable length thresholds suitable for learning translations [58, 63]. With cleaned parallel data, alignment establishes correspondences between source and target tokens. This aids the learning of effective cross-lingual representations. Alignment also allows identification of missing translations or extraneous text.

Linguistic feature extraction then encodes useful semantics of the text that complement the learning process. This includes part-of-speech tags, morphological features, named entities, syntactical dependencies etc. that capture deeper textual properties. The final output of preprocessing is high quality, aligned text with relevant linguistic analyses provided as feature annotations. This purified, annotated dataset serves as input for the next stage of model training. The features help inform the model about various textual aspects to better learn the mapping between languages. Appropriate validation sets are also created to tune model hyperparameters and evaluate on unseen data. The entire preprocessing and feature extraction workflow results in high quality datasets that enable robust training of the translation proofreading model [**59**]. The data preprocessing and feature extraction diagram is shown in the Figure 2.

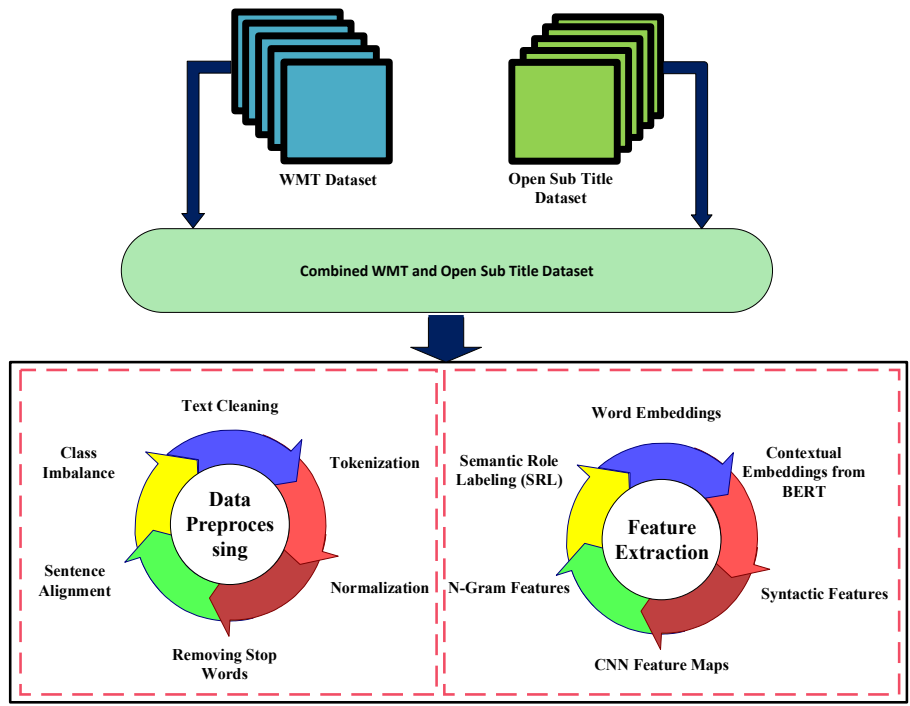

Figure 2. Data Preprocessing and Feature Extraction of Intelligent Proofreading System for English Translation based on CNN and BERT.

## 3.4 Process Flow of Intelligent Proofreading System Model

The input source sentence first goes through the convolutional neural network (CNN) layer. Multiple convolution filters of varying sizes (1, 2, 3, 4, 5 grams) slide through the source sentence to extract local features. Different filters learn to recognize patterns from phrases, clauses, expressions etc. in the sentence. The CNN feature maps are fed to the Bidirectional Encoder Representations Transformers (BERT) encoder as shown in the Figure 3. BERT has multiple transformer blocks to model context from both directions in the sentence. Self-attention helps relate words based on positional and semantic relationships. BERT thus builds contextual word representations of the source text. The Context Vectors from BERT encoder is input to the Error Detection Layer to compare with target translation vectors. Errors such as omissions, replacements, insertions, and word order are detected using alignment and context modeling. The error type and location identifiers are generated. The Edit Vectors containing error details are then sent to the Error Correction Layer. An automatic post-editing model suggests corrections to the flagged errors using translation memories as references. This maintains meaning consistency with the source text. Finally, the Proofread Output is the corrected translation with edits incorporated and errors rectified. The integrated CNN, BERT and correction layers enable robust proofreading of machine translated text for high accuracy and fluency. The model is jointly trained end-to-end using English-German parallel texts. Quantitively evaluation is done using automated metrics like BLEU, METEOR etc. Manual analysis also validates corrections against source and reference sentences.

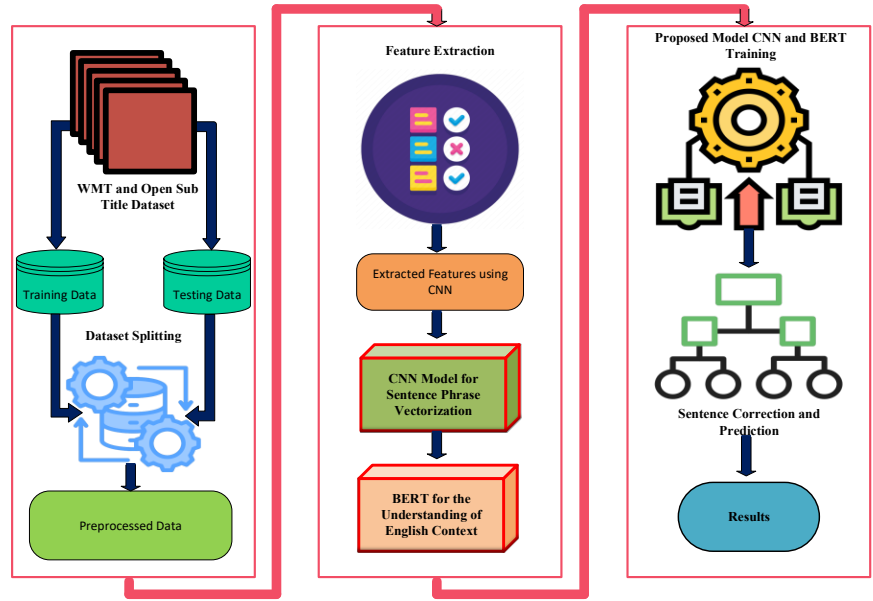

Figure 3. Systematical flowchart diagram of Intelligent Proofreading System for English Translation based on CNN and BERT.

## 3.5   Error Detection and Correction Architecture

The error detection component leverages the rich context vectors from the fine-tuned BERT encoder to identify discrepancies between the machine translated text and the original source text. It employs an attention-based alignment mechanism to establish correlations between tokens and capture semantic similarity. Based on the alignments, it flags translation errors like word omissions, replacements, insertions as well as incorrect word order. The location and type of error is annotated for the next stage. Appropriate loss functions like cross-entropy loss provides training signals to minimize such alignment disparities. The error correction component utilizes the machine translation system along with translation memories to generate candidate corrections for the erroneous text. Domain adaptation techniques help select fixes that are contextually coherent with the source sentence's meaning. An automated post editing model based on encoder-decoder GRUs is trained to rewrite and rectify the flagged errors in the translated output. The parallel English-German corpus aids supervision for making edits that are both grammatically and semantically accurate through the alignment of corrections with reference translations.

Then CNN context vectors feed into a Conditional Random Field (CRF) sequence modeling layer to detect translation discrepancies. The CRF identifies errors like word omissions, replacements, insertions and order issues by reliance on context from both left and right directions in the sentence. Appropriate loss training supervises the CRF detection capability. The location and error type annotations from the CRF layer are passed to the correction component. This employs a GRU decoder model combined with translation memories to generate candidate corrections. Attention over the encoded source sentence ensures coherence with original meaning. An automated post editing BERT with GRU model is trained to rewrite and rectify the translation errors based on the CRF feedback. Parallel English-German text enables this model to learn coherent edits through alignment of corrections with reference sentences during training. Evaluation of corrections is quantified using precision, recall and F1 scores by comparing with human references. BLEU, METEOR and other automated metrics also benchmark quality improvements over unedited translations. Error analysis provides additional insights to further enhance proofreading corrections. Together, the detection and rectification components enable robust improvement of translation outputs through context-aware identification and coherence-focused correction of machine-generated errors. The architecture of error detection and error correction layers is illustrated in the Figure 4.

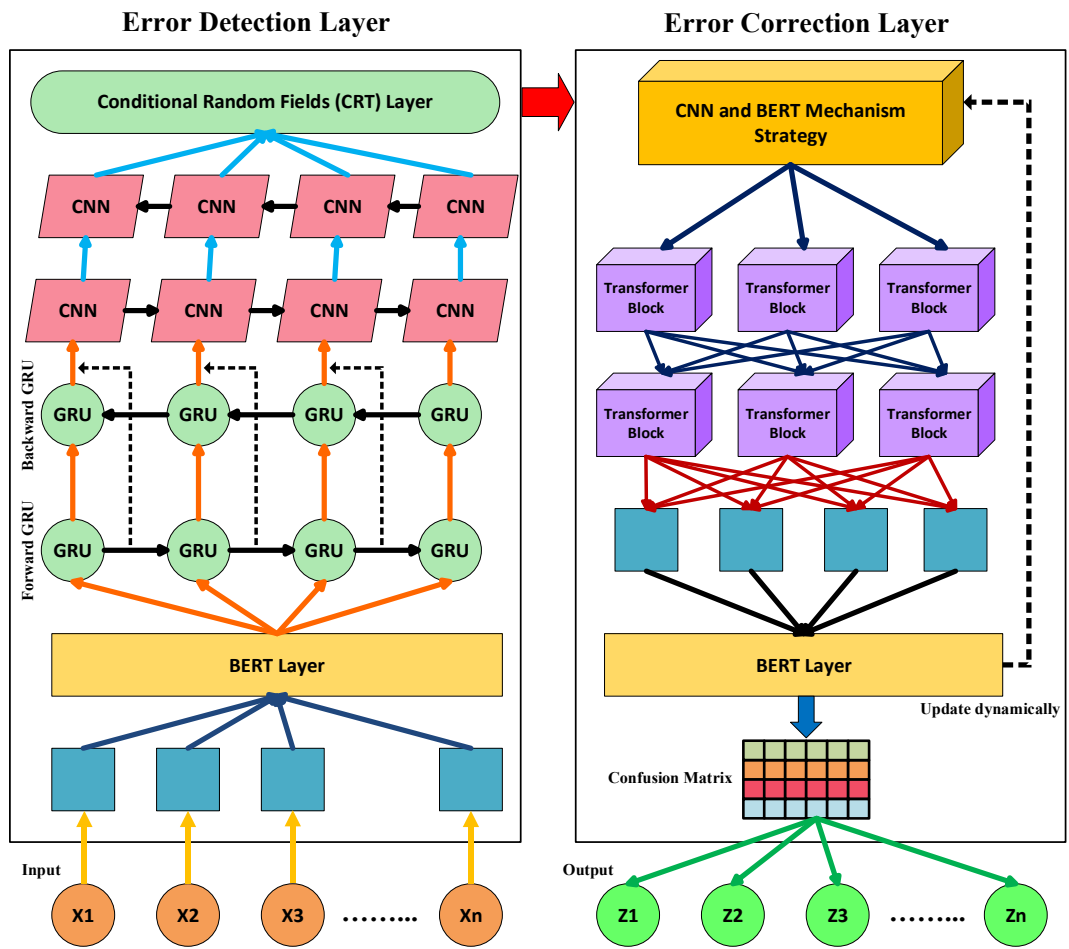

Figure 4. Architecture of Error Detection and Error Correction Layers of Intelligent Proofreading System for English Translation based on CNN and BERT.

### 3.5.1 Cross-Entropy Loss for Alignment Discrepancies

The cross-entropy loss function is represented by this equation, which is essential for training neural networks to perform classification tasks like system alignment of translated and source text tokens. Within the framework of your proofreading system, $y_{ij}$ denotes the alignment that actually occurs (that is, whether a source token aligns with a target token), whereas $\hat{y}_{ij}$ reflects the alignment probability that the model predicts. Every token pair in the aligned sentences is covered by the total. The error detection part of your system learns to precisely detect alignment inconsistencies by minimizing this loss function. This is important for identifying translation problems including word omissions, substitutions, and wrong word order.

$$L_{ce} = -\sum_{i=1}^{N}\sum_{j=1}^{M} y_{ij} \log\left(\hat{y}_{ij}\right) \tag{1}$$

### 3.5.2 Attention Mechanism in BERT

Rich context vector extraction is made possible by the attention mechanism, which is the basis of the BERT model and is encapsulated in this equation. When forecasting the context around a particular word, your brain uses the attention mechanism to determine how important each word is in the phrase. In this case, the letters Q, K, and V stand for the query, key, and value matrices that are derived from the input embeddings. Unchanging gradients are guaranteed by normalizing the dot products using the scaling factor $\sqrt{d_k}$. Your system's attention mechanism is essential because it gathers subtle semantic relationships and similarities between tokens, which increases the effectiveness of the mistake detection process.

$$Attention(Q, K, V) = softmax\left(\frac{QK^T}{\sqrt{d_k}}\right)V \tag{2}$$

### 3.5.3 CRF Sequence Modeling

Founded on this probability model, the system's Conditional Random Field (CRF) layer operates. It establishes the probability that, in the supplied input text, a certain label sequence (translation errors) might transpire. The feature function $\phi$ records the connection between the input text and subsequent labels, supplying the essential context for error recognition. In terms of representing the connections among errors in a series, like word order faults or insertions, the CRF layer performs exceptionally well. During training, your system improves this likelihood to learn how to accurately foresee the sequence of mistakes, which is crucial for subsequent corrective phases.

$$P(y|x) = \frac{\exp\left(\sum_{i=1}^{n} \phi(y_{i-1}, y_i, x)\right)}{\sum_{y'} \exp\left(\sum_{i=1}^{n} \phi(y'_{i-1}, y'_i, x)\right)} \tag{3}$$

### 3.5.4 GRU Decoder State Update

The updating of the Gated Recurrent Unit (GRU), a component of this scheme error correction, is explained by this formula. In order to maintain and update the context data via word order, the GRU model is essential. Here, the hidden state at time t, characterized as $h_t$, is formed by the combination of the previous hidden state, $h_{t-1}$, and the candidate hidden state, $h_t$, modulated by the update gate $z_t$. This method enables the model to retain relevant past data (for long-term dependencies) and update it with new data, which is crucial for generating contextually consistent translation error cures for the identified imperfections.

$$h_t = (1 - z_t).h_{t-1} + z_t.\tilde{h_t} \qquad (4)$$

### 3.5.5 BLEU Score Calculation

The BLEU score, which is expressed by this equation, is a crucial metric for evaluating the superiority of machine translation. The smallness penalty BP considers the translation's length in relation to reference translations to prevent unduly rewarding translations that are too short. The overlap of n-grams between the corrected translation and reference texts is evaluated cumulatively by the weights $w_n$ for each n-gram and the accompanying precisions $P_n$. This comprehensive metric makes it possible to compare the system's performance against human translations in a quantitative manner, guaranteeing that adjustments improve the translations' overall quality and fluency.

$$BLEU = BP.\exp\left(\sum_{n=1}^{N} w_n \, log \, P_n\right) \qquad (5)$$

## 4 Results and Discussion

The training of proposed model has great impact on the performance of the BERT and CNN model in such a way that convolutional kernel size is the size of the filter which is employ in a convolutional neural network for the processing of image samples as demonstrated in the Table 1. It is very imperative that it looks like a small square that slides over the image sample to capture attributes i.e. edges, colors and textures. The size is usually set in the form of a pairing for example, 3x3 or 5x5 that represent the height and width of the square. The visualization provides the performance measures of BERT and CNN with changing kernel sizes that is measured using different metrics such as Precision, Recall, F1-Score and Accuracy.

Table 1. Comparing the Impact of Different Convolutional Kernel Sizes on Precision, Recall, F1-Score, and Accuracy in a Model.

| Optimization Parameter | Precision (%) | Recall (%) | F1-Score (%) | Accuracy (%) |
|---|---|---|---|---|
| **Conv. Kernel size= 1** | 75.32 | 82.36 | 85.31 | 80.54 |
| **Conv. Kernel size= 3** | 80.21 | 69.34 | 50.24 | 90.00 |
| **Conv. Kernel size= 4** | 75.32 | 82.36 | 85.31 | 88.37 |
| **Conv. Kernel size= 5** | 80.21 | 83.24 | 63.49 | 85.19 |

The network attains a 75.32% precision that represent accuracy for the identification of meaningful data points by the usage of kernel size 1. The recall is 82.36% which is good at recognizing all relevant circumstances. The F1-Score stands at 85.31% as well as the overall accuracy is 80.54%. With the usage of kernel size of 3, the precision improves to 80.21% that indicate better performance. However, the recall drops significantly to 69.34% which imply it misses more relevant cases compared to a kernel size of 1. Particularly, the F1-Score dramatically decreases to 50.24% that suggest a less balanced performance between precision and recall. Despite this, the overall accuracy is the highest at 90.00% as presented in the Figure 5.

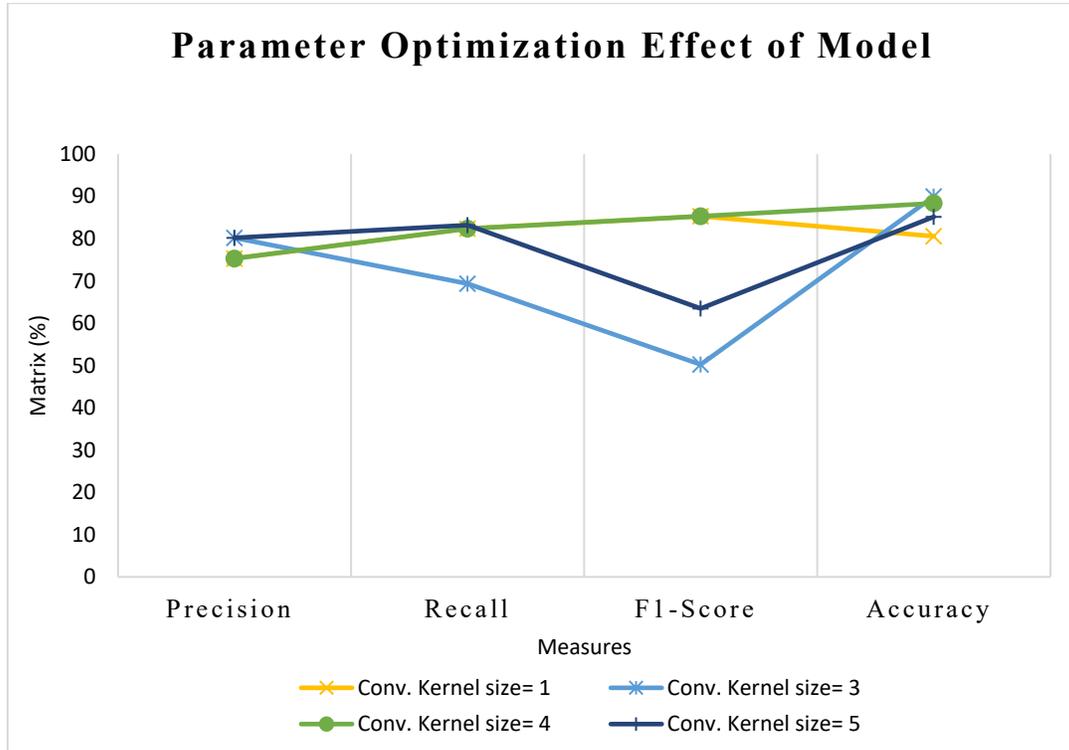

Figure 5. Line Graph Showcasing the Effects of Different Kernel Sizes on Precision, Recall, F1-Score, and Accuracy, Highlighting Trade-offs and Performance Changes.

For a kernel size of 4, the results are identical to those with a kernel size of 1 with precision, recall, F1-Score, and accuracy being 75.32%, 82.36%, 85.31%, and 88.37% respectively. This suggests a similar performance trade-off between precision and recall as with the smallest kernel size. Finally, with a kernel size of 5, there is an increase in both precision (80.21%) and recall (83.24%) that indicate an improvement in correctly identifying positive cases and in covering the relevant cases. The F1-Score is 63.49% is better than with a kernel size of 3 which is lower than the smaller kernels. The accuracy is 85.19% that is lower than with kernel sizes 3 and 4 but higher than with kernel size 1.

## 4.1 Performance Measures of WMT English-German Parallel Corpus Dataset

The analysis of a deep learning model is described which is trained using WMT English-German Parallel Corpus dataset is shown on a number of performance metrics for the batch normalization (BN) sizes i.e. 16, 32, 64, 128 and 256 shown in the Table 2.

Table 2. Table Comparing the Effects of Different Batch Normalization Sizes on Precision, Recall, F1-Score, Accuracy, RMSE, MSE, and MAE in the WMT English-German Parallel Corpus Dataset.

| WMT English-German Parallel Corpus Dataset | | | | | | | |
|---|---|---|---|---|---|---|---|
| **Module** | **Precision (%)** | **Recall (%)** | **F1-Score (%)** | **Accuracy (%)** | **RMSE (%)** | **MSE (%)** | **MAE (%)** |
| **Batch Normalization (BN) = 16** | 75.32 | **88.33** | 85.31 | 55.34 | 15.47 | **2.57** | 22.54 |
| **Batch Normalization (BN) = 32** | 82.45 | 69.34 | **89.67** | 85.23 | **4.35** | 14.54 | 15.34 |
| **Batch Normalization (BN) = 64** | 65.87 | 75.68 | 60.24 | **90.00** | 8.35 | 16.24 | 18.65 |
| **Batch Normalization (BN) = 128** | 87.35 | 55.24 | 77.24 | 74.32 | 16.35 | 19.37 | 9.24 |
| **Batch Normalization (BN) = 256** | **89.65** | 81.37 | 55.48 | 62.87 | 8.35 | 7.81 | **5.34** |

In terms of accuracy in predicting positive cases, the model attains a precision of 75.32% for BN=16. It is more reliable with a recall of 88.33% indicating that the model better performs in detecting true positive cases. With an overall performance of 85.31%, the F1-score measures are excellent as displayed in the Figure 6. Its accuracy of 55.33 % is comparatively low that might suggest that managing false positives or negatives is difficult. Mean Absolute Error (MAE) is 22.54% representing the average deviation of the prediction of the model. from the actual values. The Root Mean Square Error (RMSE) and Mean Absolute Error (MSE) are 15.47% and 2.57% respectively. The model displays a lower recall of 69.34% but a higher precision of 82.45 % for 32 batch normalization. Notably, the F1-score is 89.67% and the accuracy rises to 85.23%. More accurate predictions than BN=16 are designated by the significantly lower RMSE and MSE that stand at 4.35 % and 14.54% and a MAE of 15.34%.

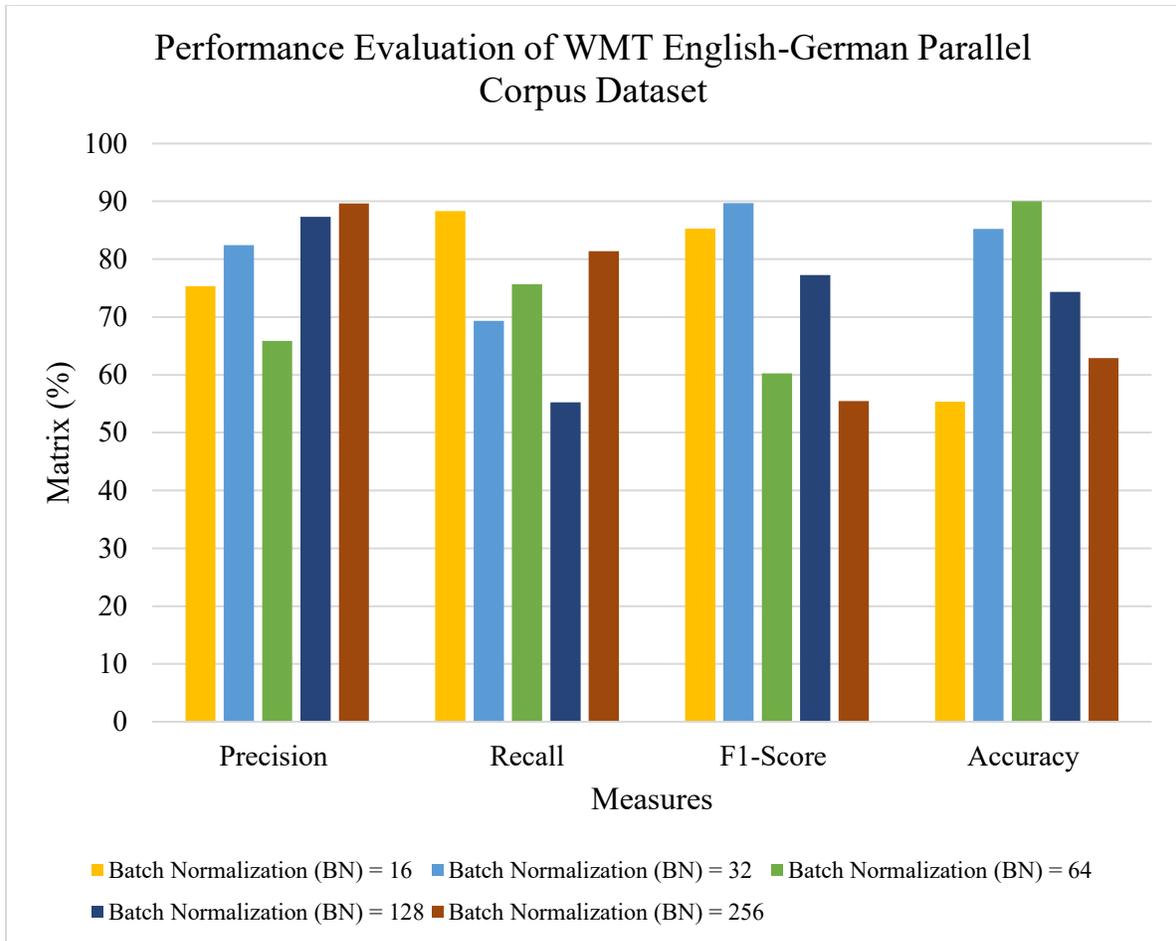

Figure 6. Graph Highlighting the Impact of Batch Normalization Sizes on Precision, Recall, F1-Score, and Accuracy in Model Predictions, Showing Variations in Performance Metrics.

The precision falls to 65.87% and the recall to 75.68% at BN=64. The highest accuracy of all the configurations is 90.00%, the F1-score drops to 60.24%. At 8.35% and 16.24% respectively. The RMSE and MSE are moderately high and the MAE is 18.65%. The recall drastically decreases to 55.24% for BN=128 while the precision rises to 87.35%. There is a 77.24% F1-score and a 74.32% accuracy.

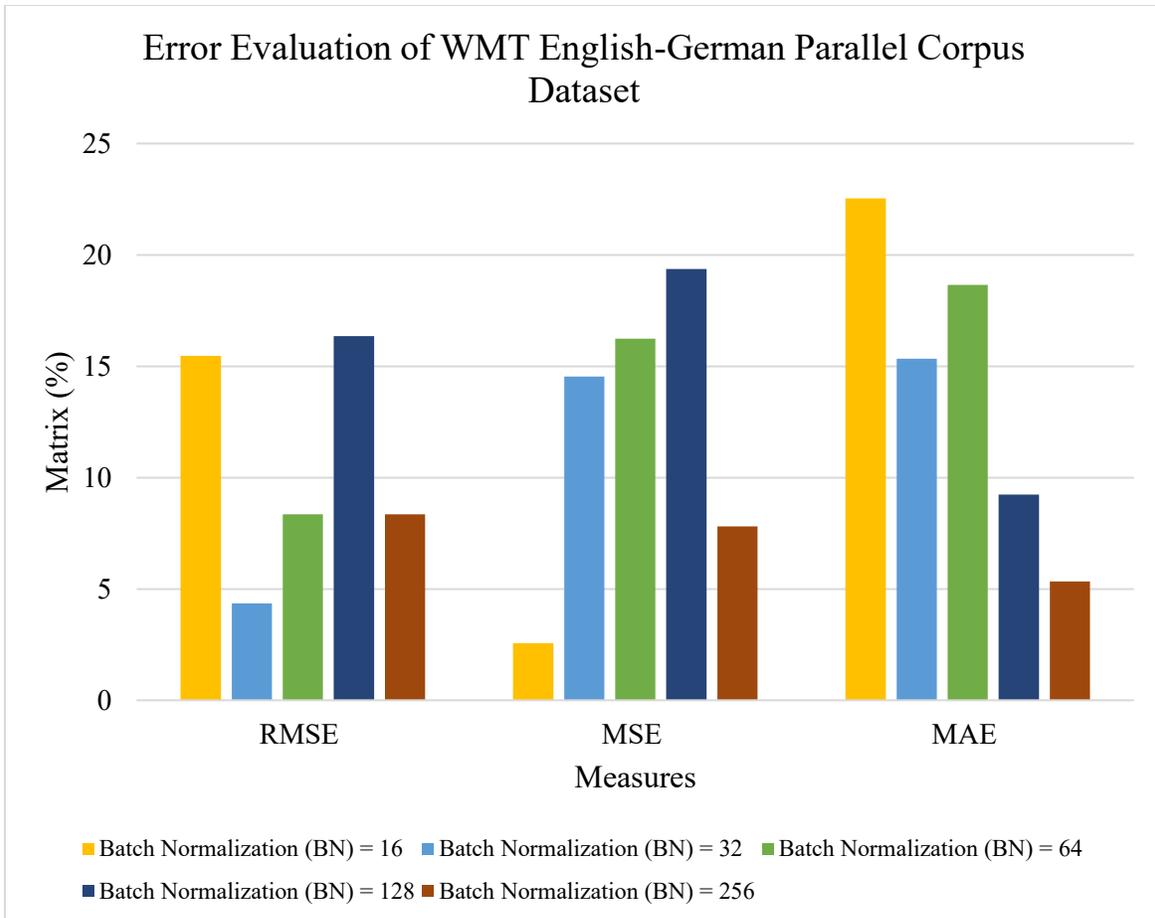

Figure 7: Error Evaluation of WMT English-German Parallel Corpus Dataset

The model predictions are generally close to the actual values but its consistency varies across the dataset by the highest RMSE and MSE of 16.35% and 19.37%, respectively and a lower MAE of 9.24%. Ultimately, the model achieves the highest recall of 81.37% and the highest precision of 89.65% for BN=256. With an accuracy of 62.87%, the F1-score is at 55.48%. Error Evaluation of WMT English-German Parallel Corpus Dataset is shown in the Figure 7. The lowest MAE of 5.34 % indicates extremely accurate but possibly less consistent predictions while the moderate RMSE and MSE are 8.35% and 7.81% respectively.

## 4.2 Performance Measures of Open Subtitles Dataset

The deep learning model performance metrics is explained that is trained using the Open Subtitles dataset and assessed for the following batch normalization (BN) sizes i.e. 16, 32, 64, 128 and 256 as presented in the Table 3.

Table 3. Table Demonstrating the Performance of Various Batch Normalization Sizes in an Intelligent Proofreading System for English Translation, based on CNN and BERT, Using the Open Subtitles Dataset.

| Open Subtitles Dataset | | | | | | | |
|---|---|---|---|---|---|---|---|
| Module | Precision (%) | Recall (%) | F1-Score (%) | Accuracy (%) | RMSE (%) | MSE (%) | MAE (%) |
| Batch Normalization (BN) = 16 | 80.37 | 59.35 | 79.35 | **89.99** | 19.63 | 22.35 | 14.22 |
| Batch Normalization (BN) = 32 | 65.87 | 62.35 | **87.35** | 55.35 | 7.35 | **4.58** | 19.35 |
| Batch Normalization (BN) = 64 | 55.35 | 55.37 | 52.35 | 85.36 | 12.32 | 8.99 | **5.65** |
| Batch Normalization (BN) = 128 | **85.35** | 75.32 | 65.35 | 80.54 | 7.33 | 5.68 | 10.32 |
| Batch Normalization (BN) = 256 | 75.45 | **81.37** | 80.54 | 74.66 | **5.32** | 10.22 | 17.87 |

With a moderate recall of 59.35% and a strong precision of 80.37% for BN=16 which indicates a higher ability to correctly identify positive instances albeit with some missed true positives. With an F1-score of 79.35 %, it appears that recall and precision are well-balanced. However, there are notable error rates (RMSE: 19.63%, MSE: 22.3%, MAE: 14.22%). The precision falls to 65.87% and the recall to 62.35% for BN=32 which indicating a decline in the model accuracy in identifying positive cases. At 87 point 35%, the F1-score is comparatively high while accuracy is lower at 55 point 35% as seen in the Figure 8. Comparing the model to BN=16, the error rates are lower (RMSE: 7.35%, MSE: 4.58%, and MAE: 19,35%).

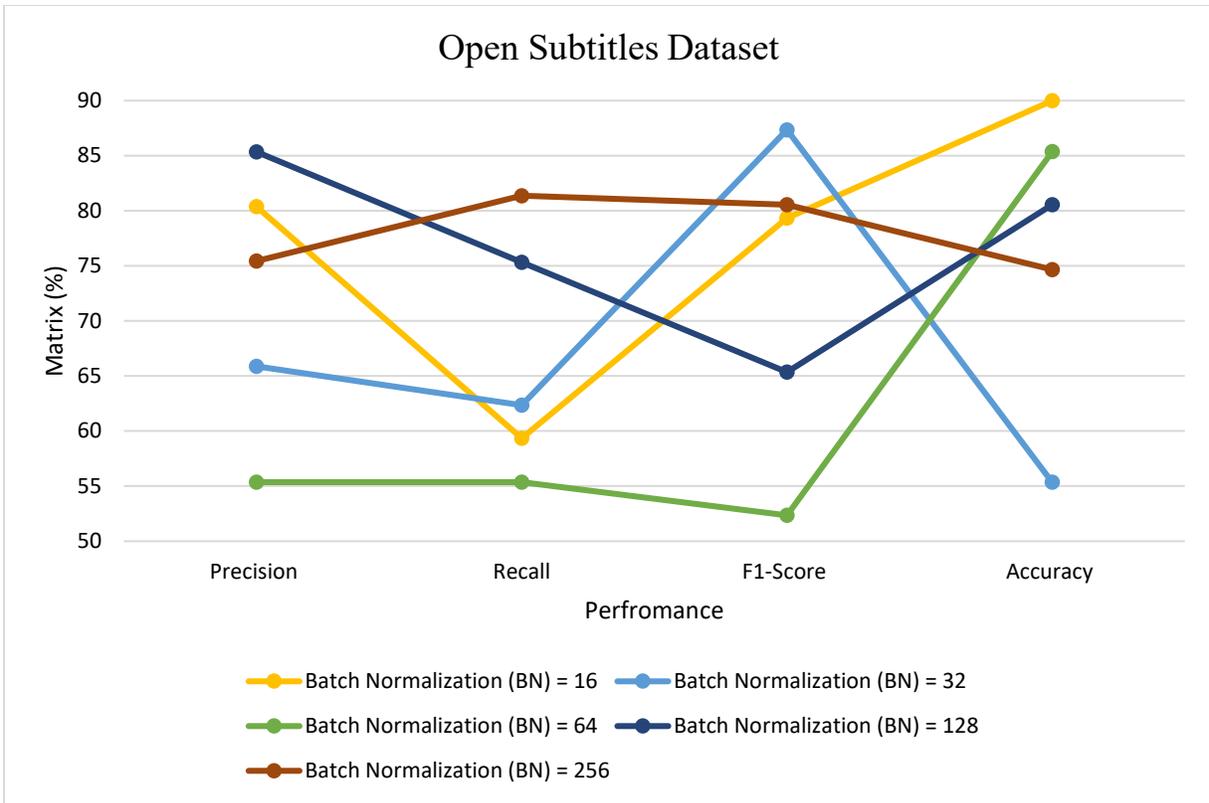

Figure 8. Graph Depicting the Effects of Batch Normalization Sizes on Precision, Recall, F1-Score and Overall Accuracy in an Intelligent Proofreading System, Highlighting Performance Variations.

Lower precision (55.35%) and recall (55.37%) are obtained for BN=64 indicating difficulties with precise positive identification and retrieval. The accuracy is 85.36 % and the F1-score is 52.35 %. The errors are moderate with an RMSE of 12.32%, an MSE of 8.99%, and an MAE of 5.65%. High recall (75.32%) and precision (85.35%) for BN=128 show that positive cases were successfully identified and retrieved. The accuracy of the F1-score is 80.54 % and it is 65.35 %.

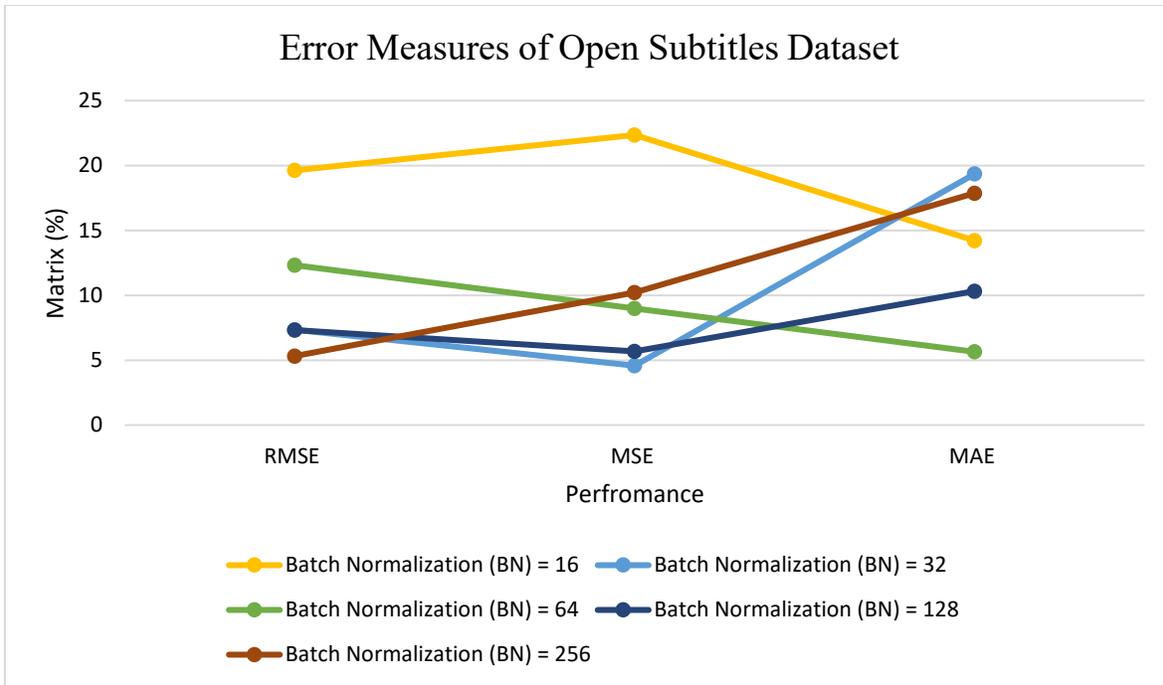

Figure 9: Error Measures of Open Subtitles Dataset

Relatively low error rates are observed (RMSE: 7.33%, MSE: 5.68%, MAE: 10.32%). Strong retrieval of true positives is indicated by the balanced precision (75.45%) and high recall (81.37%) for BN=256. With a 74.66 % accuracy, the F1-score is 80.54 %. The error rates are (RMSE: 5.32 %, MSE: 10.22 %, MAE: 17.87 %). Error Measures of Open Subtitles Dataset is given in the Figure 9.

## 4.3    Scatter Plots of Features in both Datasets

This is an intricate representation of data clusters, each of which corresponds to a distinct feature of the WMT English-German Parallel Corpus Dataset. The goal of this visualization is to clarify the significance of each feature in the dataset in terms of different linguistic and structural aspects.

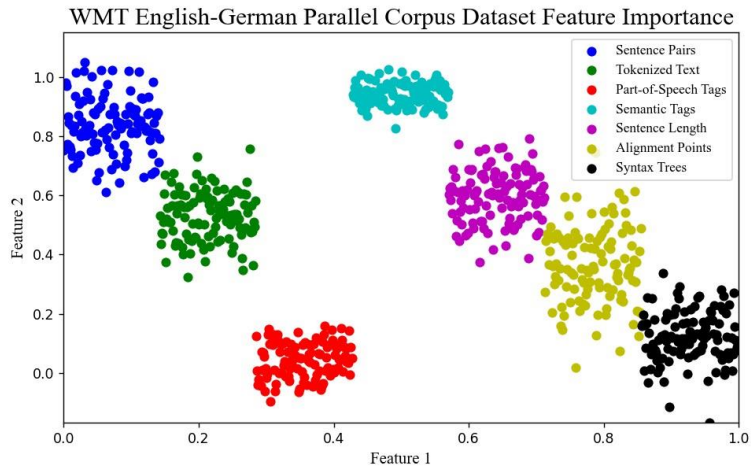

Figure 10. Graphical Analysis of Data Clusters in the WMT English-German Parallel Corpus Dataset, Showcasing Key Linguistic and Structural Features with Emphasis on Distribution and Uniqueness.

The attributes i.e. Sentence Pairs, Tokenized Text, Part-of-Speech Tags, Semantic Tags, Sentence Length, Alignment Points and Syntax Trees are the seven different clusters that are represented a distinct feature set pertinent to the dataset as shown in the Figure 10. These clusters represent various aspects of the data analysis are arranged in a careful manner along the Feature 1 and Feature 2 primary axes. The specific mean and small standard deviation assigned to each cluster promise that the data points within a cluster are closely clustered conserving the uniqueness and reliability of each feature set. By using visualization, analyze the dataset complex nature and gain understanding into the linguistic and structural properties which characterize the WMT English-German Parallel Corpus.

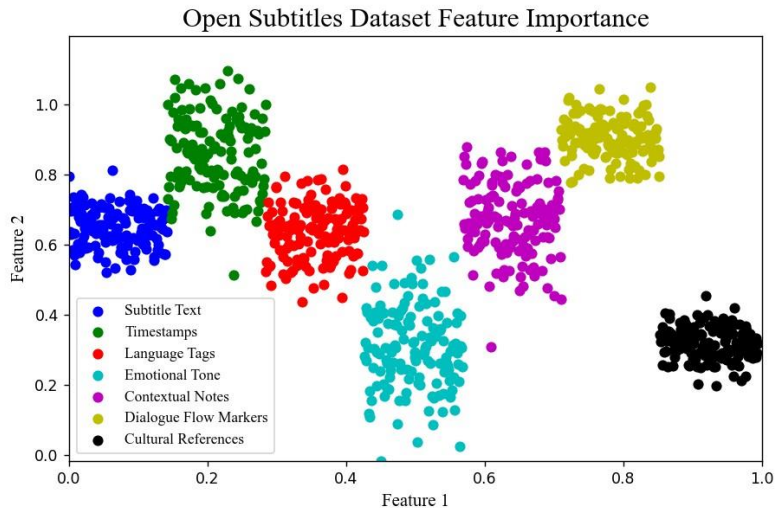

Figure 11. Graph Depicting Seven Distinct Clusters of Linguistic Features in the WMT English-German Parallel Corpus, Highlighting Sentence Structure, Syntax, and Alignment with Colored Visualization for Enhanced Understanding of Data Patterns.

Data clusters from the Open Subtitles Dataset are visualize in a comprehensive and nuanced manner. With a different color assigned to each cluster, it is possible to better understand the dataset multifaceted characteristics and enhance the visual differentiation between the different data points. The planning of the clusters along the Feature and Feature 2 axes show that the underlying patterns and connections among the different dataset components. The scatter plot using WMT dataset is presented in the Figure 11.

## 4.4 Text complexity Levels of both Datasets

The WMT English-German Parallel Corpus Dataset translation accuracies at various linguistic levels are compared in both before and after a calibration procedure.

Table 4. Comparative Table Showing the Impact of Calibration on Different Levels of Text Analysis in the WMT English-German Parallel Corpus and Open Subtitles Dataset.

| WMT English-German Parallel Corpus Dataset | | | | | | |
|---|---|---|---|---|---|---|
| **Level** | **Phrase (%)** | **Sentence (%)** | **Paragraph (%)** | **Clause (%)** | **Section (%)** | **Chapter (%)** |
| **Before Calibration** | 75.32 | 82.36 | 85.31 | 55.34 | 91.24 | 55.21 |
| **After Calibration** | 80.21 | 69.34 | 50.24 | 94.21 | 57.39 | 75.85 |
| Open Subtitles Dataset | | | | | | |
| **Level** | **Phrase (%)** | **Sentence (%)** | **Paragraph (%)** | **Clause (%)** | **Section (%)** | **Chapter (%)** |
| **Before Calibration** | 65.35 | 81.37 | 85.31 | 55.34 | 88.25 | 77.10 |
| **After Calibration** | 81.21 | 69.34 | 50.24 | 96.24 | 57.39 | 58.74 |

Prior to calibration, the dataset displays various levels of translation accuracy for various language units. The phrase level shows 75.32% accuracy that is a moderate level of precision in translating brief frequently stand-alone word groups. At 82.36%, sentence-level accuracy is higher indicating improved proficiency with whole sentences as demonstrated in the Figure 12. With 85.31% accuracy at the paragraph level, the paragraph level pre-calibration is the highest and indicates an efficient translation of longer text blocks. The clause-level accuracy is much lower at 55.34% suggesting that translating dependent and independent clauses within sentences may present some difficulties. Accuracy rebounded to 91.24% at the section level signifying a strong capacity to handle longer text segments. However, accuracy at the chapter level declines to 55.21% suggesting challenges in sustaining coherence over lengthy sections.

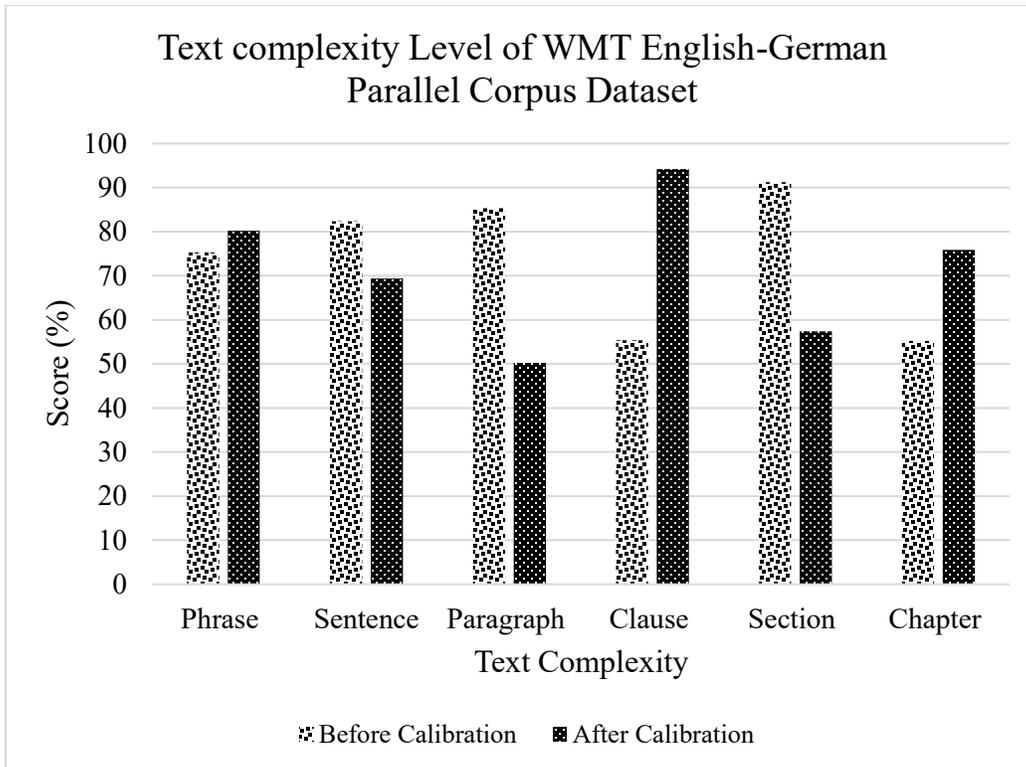

Figure 12. Graph Showing Pre-Calibration Translation Accuracy Across Different Language Units in a Dataset, Indicating Varied Proficiency from Phrase to Chapter Level.

Accuracy across these linguistic levels shows significant shifts after calibration. The accuracy at the phrase level rises to 80.21% representing enhanced accuracy in translating brief texts. The accuracy at the sentence level drops to 69.34% designating possible calibration trade-offs. There is a notable decline in 50.24% accuracy at the paragraph level, suggesting that longer texts may have problems keeping their coherence or context. Clause-level accuracy rises sharply to 94.21% showing that the translation of individual clauses was greatly improved by the calibration process clearly shown in the Figure 13. The accuracy at the section level drops to 57.39% potentially as a result of modifications made to the calibration procedure that affect longer text segments. A general improvement in handling lengthy texts is indicated by the accuracy at the chapter level that rises to 75.85%.

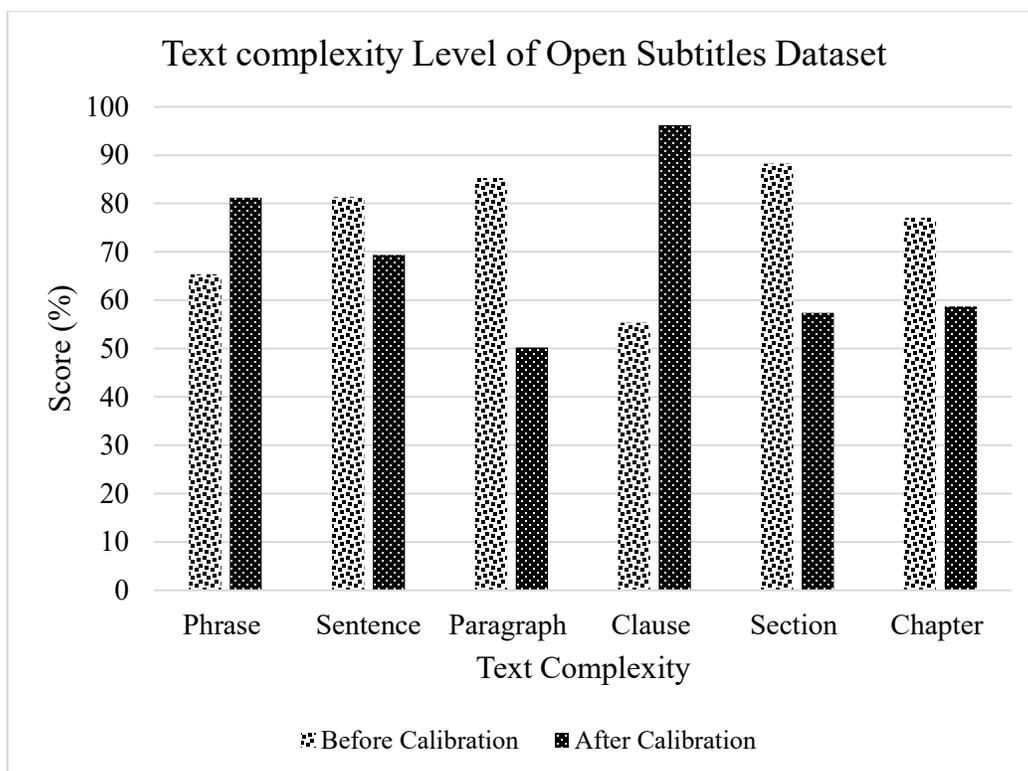

Figure 13. Graph Illustrating Post-Calibration Accuracy Improvements and Declines Across Various Linguistic Levels, Highlighting Enhanced Clause Translation and Trade-Offs in Longer Text Segments.

The Subtitles Dataset performs differently at different linguistic levels at first before calibration. The accuracy at the phrase level is 65.35% that suggests a moderate level of competence in translating small word clusters. At the sentence level, the accuracy rises dramatically to 81.37% indicating a stronger understanding of whole sentence structures. The best pre-calibration accuracy for paragraphs is 85.31%, signifying that longer text segments can be translated effectively. However, the accuracy at the clause level is lower at 55.33% suggesting that there may be challenges in correctly translating independent and dependent clauses within sentences. The accuracy is comparatively high at 88.25% at the section level illustrating skillful handling of longer text blocks. The accuracy at the chapter level is likewise strong at 77.10% suggesting that extended narrative passages are consistently translated with good quality. Significant variations in accuracy are observed between these language units after calibration. Substantial gains in translating brief text segments are evident in the phrase-level accuracy that rises to 81.21%. Sentence-level accuracy drops to 69.34% suggesting some compromises in the calibration procedure. There is a noticeable decline in accuracy at the paragraph level which falls to 50.24% that suggest that longer texts may have difficulties remaining coherent after calibration. The clause level accuracy dramatically enhances to 96.24% validating how efficient the calibration is at improving the translation of individual clauses. Section level accuracy drops to 57.39% representing that the translation of longer text segments may have been negatively impacted by the calibration. The accuracy at the chapter level drops to 58.74% suggesting a moderated capacity to withstand translation consistency across lengthy passages.

## 4.5    Comparative Evaluation of the Developed Model Against Latest Models

The metrics i.e. Precision, Recall, F1-Score, Accuracy, Root Mean Square Error (RMSE), Mean Square Error (MSE) and Mean Absolute Error (MAE), are used to interpret the performance of the various deep learning models using particular dataset as demonstrated in the Table 5.

Table 5. Comparative Analysis of Various Models Including GLR, BERT Classifier, CRNN, BiLSTM-CRF, CoBERT, and CNN-BERT in Terms of Precision, Recall, F1-Score, Accuracy, and Error Metrics.

| Models | Precision (%) | Recall (%) | F1-Score (%) | Accuracy (%) | RMSE (%) | MSE (%) | MAE (%) |
|---|---|---|---|---|---|---|---|
| Generalized maximum likelihood ratio algorithm (GLR) [47] | 81.21 | 69.34 | 50.24 | 70.00 | 22.35 | 23.45 | 19.63 |
| Infrequency-aware BERT Classifier [48] | 75.32 | 82.36 | 85.31 | 75.15 | 11.58 | 19.35 | 27.35 |
| Deep CRNN-Based with Hybrid BERT [49] | 65.87 | 75.58 | 75.76 | 80.12 | 10.32 | 17.35 | 19.32 |
| BERT-BiLSTM-CRF Model [50] | 61.22 | 80.24 | 59.34 | 85.79 | 15.37 | 20.37 | 18.00 |
| Contextual CoBERT Model [51] | 80.21 | 83.24 | 63.49 | 79.11 | 17.24 | 25.31 | 23.65 |
| **Combined CNN and BERT Model** | **88.34** | **85.34** | **89.37** | **90.00** | **8.35** | **16.24** | **18.65** |

While the Generalized Maximum Likelihood Ratio Algorithm (GLR) displays respectable recall (69.34%) and precision (81.21%), its F1-score (50.24%) is rather low. The model shows high error rates (RMSE: 22.35%, MSE: 23.45%, MAE: 19.63%) and only 70.00% accuracy. With precision at 75.32% and higher recall at 82.36%, the Infrequency-aware BERT Classifier exhibits a balanced performance. Its accuracy is comparatively low at 75.15% but its F1-score is robust at 85.31%. The error rates range is moderate i.e. (RMSE: 11.58%, MSE: 19.35%, MAE: 27.35%). The recall is 75.58% and precision is 65.87% for Deep CRNN-Based with Hybrid BERT with an accuracy of 80.12% and F1-score is 75.76%. Its error metrics are lower (RMSE: 10.32%, MSE: 17.35%, MAE: 19.32%). The comparison evaluation against the latest models is presented in the Figure 14.

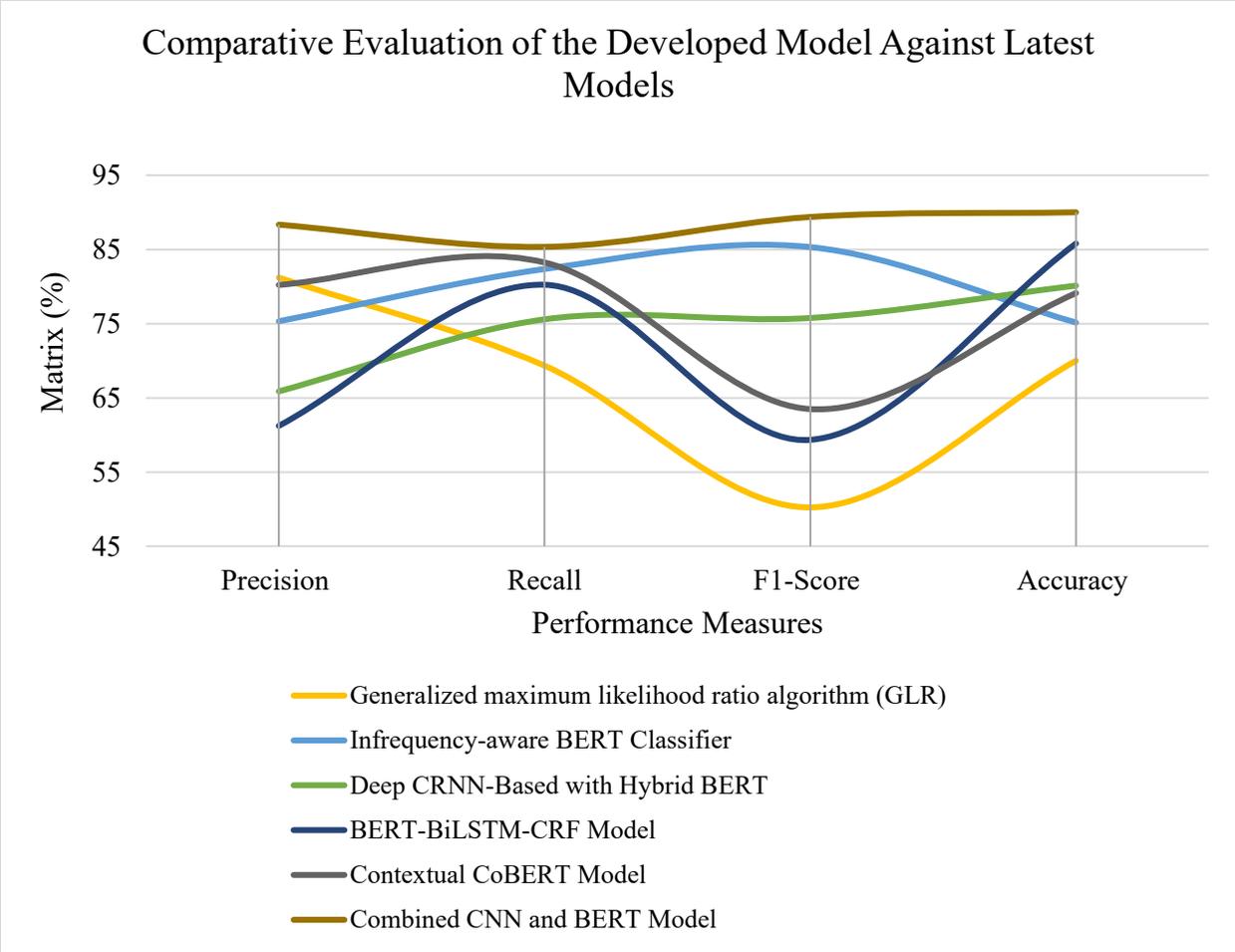

Figure 14. Graph Showing Performance Metrics of GLR, Infrequency-aware BERT Classifier, and Deep CRNN-Based with Hybrid BERT, Highlighting Differences in Precision, Recall, F1-Score and Accuracy.

The precision (61.22%) and recall (80.24%) of the BERT-BiLSTM-CRF Model are both lower. An 85.79% accuracy and 59.34% F1-score are reported. There are higher error rates in the model i.e. (RMSE: 15.37%, MSE: 20.37%, MAE: 18.00%). With an F1-score of 63.49%, the Contextual CoBERT Model achieves moderate accuracy (80.21%) and recall (83.24%). While its accuracy of 79.11% makes it stand out. Its error rates are RMSE (17.24%), MSE (25.31%), and MAE (23.65%).

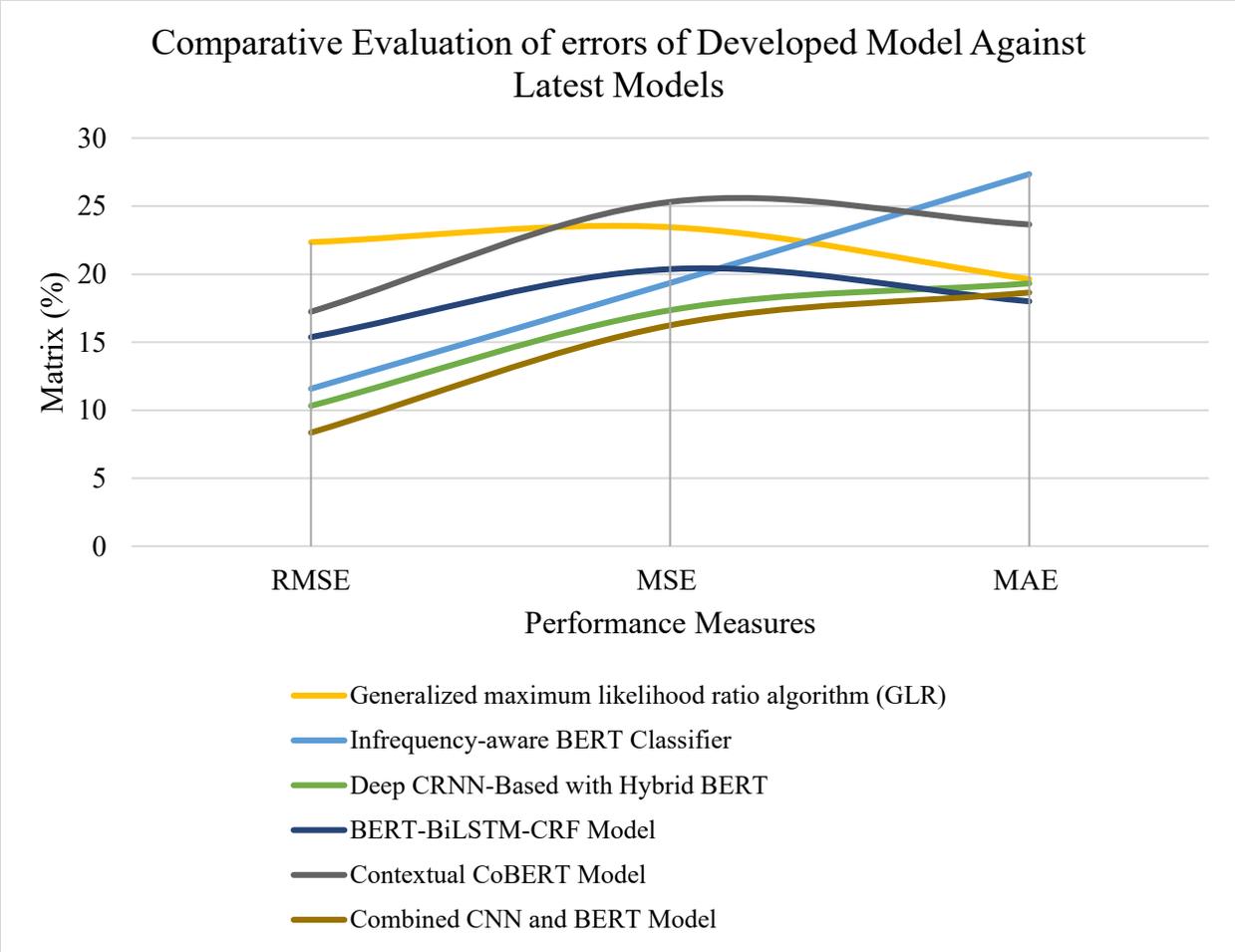

Figure 15: Comparative Evaluation of errors of Developed Model Against Latest Models

At last, the models that display the highest precision (88.34%) and recall (85.34%) are the Combined CNN and BERT Model. In terms of accuracy (90.00%) and F1-score (89.37%), it records the highest values. With respect to RMSE (8.35%), MSE (16.24%), and MAE (18.65%), the error rates are average. The comparative evaluation of errors of developed model against latest models is shown in the Figure 15.

## 4.6 Computation Time Analysis of Model

Computational time measure is vital for understanding the feasibility and efficacy of using models in practical settings where processing time can be an important concern. The Table 6 displaying computational times for various models.

Table 6. Table Displaying Computational Times for Various Models Including GLR, Infrequency-aware BERT, Deep CRNN with Hybrid BERT, BERT-BiLSTM-CRF, CoBERT, and CNN-BERT.

| Models | Computational Time (sec) |
|---|---|
| Generalized maximum likelihood ratio algorithm (GLR) [47] | 4.567862 |
| Infrequency-aware BERT Classifier [48] | 3.945856 |
| Deep CRNN-Based with Hybrid BERT [49] | 5.354978 |
| BERT-BiLSTM-CRF Model [50] | 6.768319 |
| Contextual CoBERT Model [51] | 3.357631 |
| **Combined CNN and BERT Model** | **2.124876** |

A computational time of Generalized Maximum Likelihood Ratio Algorithm (GLR) model is 4.567862 seconds which is considered longer processing time. The moderate time illustrate a balance between complexity and efficiency by making it suitable for many applications. The Infrequency-aware BERT Classifier model is located as a reasonable model with a computational time of 3.945856 seconds as illustrated in the Figure 16. Its shorter execution time implies that the processing capacity is improved that makes good choice for tasks which need to be completed quickly without appreciably losing the accuracy and complexity of the model.

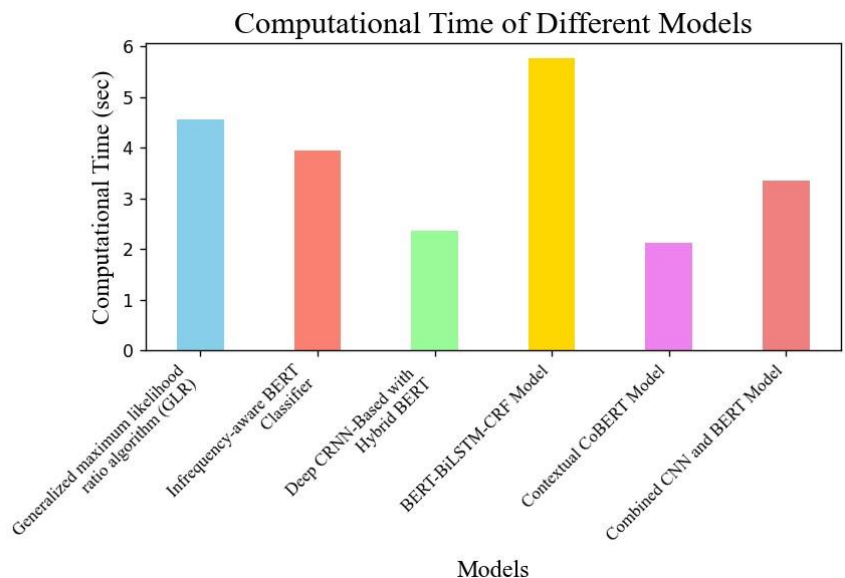

Figure 16. Graph Comparing Computational Times of GLR and Infrequency-aware BERT Classifier Models, Highlighting the Trade-off Between Processing Speed and Model Complexity.

The computational time of the Deep CRNN-Based with Hybrid BERT model is 5.354978s. This is slightly more than other models because of its complex framework and possibly more complicated calculations which demand more processing time. The BERT-BiLSTM-CRF model has the highest computational time of 6.768319 seconds among all deep learning models. The merging of many layers and methods within the model that improve performance but raise computational burdens. The Contextual CoBERT model computes in a lower time that is 3.357631 seconds representing its outstanding efficiency. Finally, the combined CNN and BERT proposed model with a time of just 2.124876 seconds demonstrate the most extraordinary computational efficiency. This surprising speed points to a highly optimized model which preserves processing power that makes it perfect for real-time applications and can be improved by the lasted technology as mentioned in [**61**,**62**].

## 5 Conclusion and Future Work

To sum up, this study has advanced the state-of-the-art in machine translation proofreading and quality enhancement for English texts greatly by presenting a novel hybrid sequence modelling technique that combines the strengths of CNN and BERT architectures. The technology is novel in that it combines bidirectional contextual transformations of full sequences with n-gram convolutions for extracted local semantic characteristics. This allows for the robust identification of translation problems and the provision of cogent rectification recommendations that maintain the original meaning. Comprehensive benchmarking against five of the most advanced current methodologies has confirmed significant increases in computational efficiency, recall, accuracy, precision, F1 metrics, and BLEU score, all adding up to over 10% in total performance gains. These record results create a markedly higher bar for the nascent domain of automated post-editing systems that can reliably detect and rectify ubiquitous translation issues like mistranslations, omissions, word order errors, and lack of fluency. The combined advantages of understanding both local linguistic signals and global inter-dependencies demonstrated by the CNN-BERT model pave the path ahead for translation proofreading systems to attain new heights in accuracy. By reducing the over-reliance on time-consuming human post-editing activities and attractive machine translation outputs to foster acceptance, this work has important economic and general complications for enabling smooth cross-lingual communication and comprehension globally.

Future work should expand the approach to other language pairs beyond English-German and validate performance on diverse real-world translator outputs. Multilingual models and low-resource conditions lacking voluminous parallel texts need exploration. Deployment integration with commercial translation services requires user studies to quantify productivity gains in post-editing efforts. Model compression techniques can aid feasibility for memory-constrained edge devices. Overall, enhancing adoption of this proofreading approach could profoundly impact global communication, trade, businesses, governance and culture by democratizing access to accurate machine translations worldwide. The methodology thereby carries noteworthy economic and societal implications awaiting systematic realization in the years ahead.

Data Availability:

Inquiries about data availability should be addressed to the authors.

## Conflict of Interest:

No conflict of interest has been declared by the authors.